%% file: sample.tex
\documentclass[twoside,11pt]{article}

\usepackage{blindtext}

\usepackage[preprint]{jmlr2e}
\usepackage{amsmath}
\usepackage{amssymb}
\usepackage{pifont}
\usepackage{listings}
\usepackage{xcolor}

\definecolor{codegreen}{rgb}{0,0.6,0}
\definecolor{codegray}{rgb}{0.5,0.5,0.5}
\definecolor{codepurple}{rgb}{0.58,0,0.82}
\definecolor{backcolour}{rgb}{0.95,0.95,0.92}

\lstdefinestyle{mystyle}{
    backgroundcolor=\color{backcolour},   
    commentstyle=\color{codegreen},
    keywordstyle=\color{blue},
    numberstyle=\tiny\color{codegray},
    stringstyle=\color{codepurple},
    basicstyle=\ttfamily\footnotesize,
    breakatwhitespace=false,         
    breaklines=true,                 
    captionpos=b,                    
    keepspaces=true,                 
    numbers=left,                    
    numbersep=5pt,                  
    showspaces=false,                
    showstringspaces=false,
    showtabs=false,                  
    tabsize=2
}

\newcommand{\cmark}{\ding{51}}%
\newcommand{\xmark}{\ding{55}}%
\newcommand{\MATLAB}{\textsc{Matlab}\xspace}
\usepackage{lastpage}
\ShortHeadings{PyXAB - A Python Library for $\mathcal{X}$-Armed Bandit and Online Blackbox Optimization}{Li, Li, Honorio and Song}
\firstpageno{1}

\begin{document}

\title{PyXAB - A Python Library for $\mathcal{X}$-Armed Bandit and Online Blackbox Optimization Algorithms}

\author{\name Wenjie Li \email li3549@purdue.edu \\
       \addr Department of Statistics,
       Purdue University
       \thanks{The two authors have contributed equally to this project }
       \AND
       \name Haoze Li \email li4456@purdue.edu \\
       \addr Department of Statistics,
       Purdue University
       \footnotemark[1]
       \AND    
        \name Jean Honorio \email jhonorio@purdue.edu \\
       \addr Department of Computer Science,
       Purdue University
       \AND
        \name Qifan Song \email qfsong@purdue.edu \\
       \addr Department of Statistics,
       Purdue University
       \AND
       }
\editor{My editor}

\maketitle

\begin{abstract}%   <- trailing '%' for backward compatibility of .sty file
We introduce a Python open-source library for $\mathcal{X}$-armed bandit and online blackbox optimization named PyXAB. PyXAB contains the implementations for more than 10 $\mathcal{X}$-armed bandit algorithms, such as \texttt{HOO}, \texttt{StoSOO}, \texttt{HCT}, and the most recent works \texttt{GPO}  and \texttt{VHCT}. PyXAB also provides the most commonly-used synthetic objectives to evaluate the performance of different algorithms and the various choices of the hierarchical partitions on the parameter space. The online documentation for PyXAB includes clear instructions for installation, straight-forward examples, detailed feature descriptions, and a complete reference of the API. PyXAB is released under the MIT license in order to encourage both academic and industrial usage. The library can be directly installed from PyPI with its source code available at \url{https://github.com/WilliamLwj/PyXAB}.
\end{abstract}


\input{section_manager.tex}

\end{document}

%% file: section_manager.tex
\input{sections/01.Introduction.tex}

\input{sections/02.API_Design.tex}
\input{sections/03.Code_Quality.tex}
\input{sections/04.Conclusions.tex}

\onecolumn
\bibliography{sample}

%% file: sections/01.Introduction.tex
\section{Introduction}
\label{sec: introduction}

Online blackbox optimization has become a heated research topic due to the recent popularity of machine learning models and thus the increasing demand for hyper-parameter tuning algorithms\citep{Li2018Hyperband, shang2019general}. Other applications, such as neural architecture search, federated learning, and personal investment portfolio designs, also contribute to its prosperity nowadays \citep{li2021optimumstatistical, Li2022Federated}. Different online blackbox optimization algorithms, e.g., Bayesian Optimization algorithms \citep{Shahriari2016Taking} and two-point evaluation methods \citep{DuchiOptimal, Shamir2015An} have been proposed.

\begin{table}
    \centering
    \caption{Selected examples of $\mathcal{X}$-armed bandit algorithms implemented in our library. \textit{Cumulative}: whether the algorithm focuses on optimizing cumulative regret or simple regret. \textit{Stochastic}: whether the algorithm deals with noisy rewards. \textit{Open-sourced?}:  the code availability before the development of PyXAB.}
    \begin{tabular}{l c c c}
        \hline
        {$\mathcal{X}$-Armed Bandit Algorithm} 
        & Cumulative & Stochastic & {Open-sourced?}   \\
        \hline
        \texttt{HOO} \citep{bubeck2011X} & \cmark & \cmark & \cmark (Python)  \\
        \texttt{DOO} \citep{Munos2011Optimistic} & \xmark & \xmark & \xmark   \\
        \texttt{StoSOO} \citep{Valko13Stochastic} & \xmark & \cmark & \cmark (\MATLAB, C)  \\
        \texttt{HCT} \citep{azar2014online} & \cmark & \cmark & \xmark   \\
        \texttt{POO} \citep{Grill2015Blackbox} & \xmark& \cmark & \cmark (Python, R) \\
        \texttt{GPO} \citep{shang2019general} & \xmark & \cmark &\xmark  \\
        \texttt{PCT}  \citep{shang2019general} & \xmark & \cmark &\xmark \\
        \texttt{SequOOL}  \citep{bartlett2019simple} & \xmark & \xmark &\xmark \\
        \texttt{StroquOOL}  \citep{bartlett2019simple} & \xmark & \cmark &\xmark \\
        \texttt{VHCT}  \citep{li2021optimumstatistical} & \cmark & \cmark &\xmark \\
       \hline
    \end{tabular}
        \vspace{-10pt}
    \label{tab: summary}
\end{table}

Apart from the aforementioned works, another very famous line of research is $\mathcal{X}$-armed bandit, also known as Lipschitz bandit, global optimization or bandit-based blackbox optimization \citep{kleinberg2008multi-armed, bubeck2011X, Grill2015Blackbox, bartlett2019simple}. In this field, researchers split the parameter domain $\mathcal{X}$ into smaller and smaller sub-domains (commonly known as nodes) hierarchically, and treat each sub-domain to be an un-evaluated arm as in the multi-armed bandit problems \citep{bubeck2011X, azar2014online}.  However, such $\mathcal{X}$-armed bandit problems are much harder than their multi-armed counterparts, since the number of sub-domains increase exponentially as the partition grows, and the hierarchical structure implies internal correlations between the ``arms". 

Despite the popularity of this area, most of the algorithms proposed by the researchers are either not open-sourced or are implemented in different programming languages in disjoint packages. For example, \texttt{StoSOO}\citep{Valko13Stochastic} is implemented in MATLAB and C\footnote{\url{https://team.inria.fr/sequel/software/}}, whereas \texttt{HOO} \citep{bubeck2011X}  is implemented in Python\footnote{\url{https://github.com/ardaegeunlu/X-armed-Bandits}}. For most of the other algorithms, no open-sourced implementations could be found on the internet. We believe the lack of such resources results from the following two main reasons. 

\begin{itemize}
    \item The algorithms are long and intrinsically hard to implement due to the heavy usage of hierarchical partitions, node sampling, and the exploration-exploitation strategies that involve building, maintaining, and expanding complicated tree structures. It could take multiple days to implement and test one single algorithm.
    
    \item The problem settings for the algorithms could be slightly different. As shown in Table \ref{tab: summary}, some algorithms such as \texttt{HOO} and \texttt{HCT} are designed for the setting where the function evaluations can be noisy, while \texttt{SequOOL} is proposed for the noiseless setting. Some algorithms focus on cumulative-regret optimization where as some only care about the last-point regret or the simple regret\footnote{ A more detailed discussion on simple regret and cumulative regret can be found in \citet{bubeck2011X}}. Therefore, experimental comparisons often focus on a small subset of algorithms, see e.g., \citet{azar2014online}, \citet{bartlett2019simple}. The unavailability of a general package only deteriorates such situation. 
\end{itemize}

To remove the barriers for future research in this area, we have developed PyXAB, a Python library of the existing popular $\mathcal{X}$-armed bandit algorithms. To the best of our knowledge, this is the first comprehensive library for $\mathcal{X}$-armed bandit, with clear documentations and user-friendly API references.

%% file: sections/02.API_Design.tex
\section{Library Design and Usage}
\label{sec: api_design}

\begin{figure}
    \vspace{-20pt}
    \centering
    \includegraphics[width=0.8\linewidth]{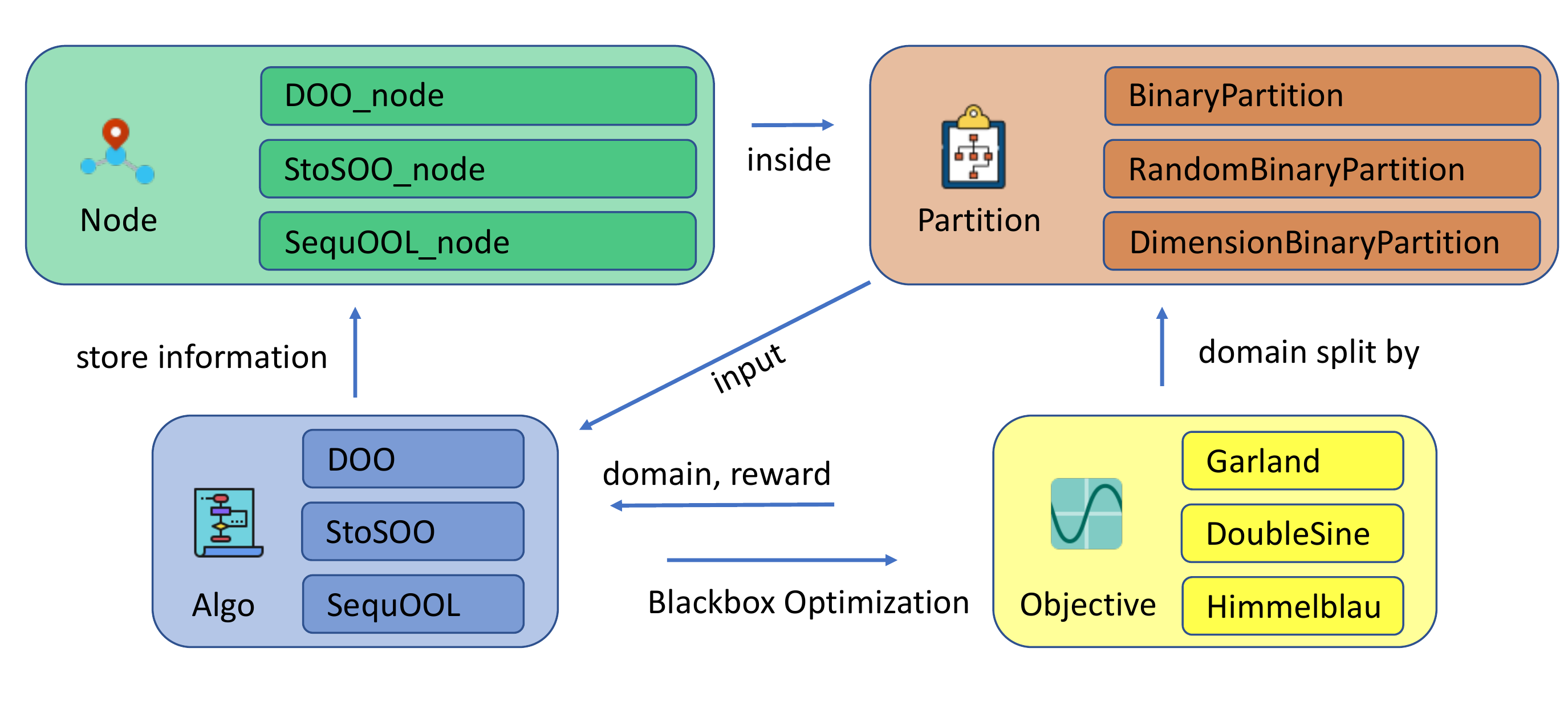}
    \vspace{-15pt}
    \caption{Overview of the PyXAB library}
    \label{fig: overview}
    \vspace{-15pt}
\end{figure}

The API of PyXAB is designed to follow the $\mathcal{X}$-armed bandit learning paradigm and to allow the maximum freedom of usage. We provide an overview of the library in Figure \ref{fig: overview}. 

\textbf{Algorithm}. All the algorithms inherit the abstract class \texttt{Algorithm}. Each algorithm is required to implement two functions: (1) a \texttt{pull()} function that returns the chosen point to be evaluated by the objective; (2) a \texttt{receive\_reward()} function to collect the evaluation result and update the algorithm behavior. 

\textbf{Partition}. Given any parameter domain, the user is able to choose any partition of the domain as part of the input of the optimization algorithm. All implemented partitions inherit the \texttt{Partition} class, which has useful base functions such as \texttt{deepen} and \texttt{get\_node\_list}. Each specific partition class needs to implement a unique \texttt{make\_children()} function that split one parent node into the children nodes and maintain the tree structure. We provide a few choices such as \texttt{BinaryPartition} and \texttt{RandomBinaryPartition}.

\textbf{Node}. The base node class used in any partition is \texttt{P\_node}, which contains useful helper functions to store domain information and maintain the partition structure. However, we allow the algorithms to overwrite the node choices in any partition so that node-wise operations are allowed. For example, the \texttt{StoSOO} algorithm needs to compute and store the $b_{h,i}$-value for each node \citep{Valko13Stochastic}. The \texttt{StroquOOL} algorithm needs to record the number of times a node is opened \citep{bartlett2019simple}. Therefore, different node classes are implemented for these algorithms.

\textbf{Objective}. For all the objectives implemented in this package, they all inherit the \texttt{Objective} class and all have a function \texttt{f()} that returns the evaluation result of a given point. We provide all commonly used synthetic objectives that evaluate the performance of $\mathcal{X}$-armed bandit algorithms in research papers, such as \texttt{Garland}, \texttt{DoubleSine}, and \texttt{Himmelblau}.

The usage of the PyXAB library is rather straight-forward. Given the number of rounds, the objective function, and the parameter domain, the learner would choose the partition of the parameter space and the bandit algorithm. Then in each round, the learner obtains one point from the algorithm, evaluate it on the objective, and return the reward to the algorithm. 
The following snippet of code provides an example of optimizing the Garland synthetic objective on the domain $[[0, 1]]$ by running the \texttt{HCT} algorithm with \texttt{BinaryPartition} for 1000 iterations. As can be observed, only about ten lines of code are needed for the learning process apart from the import statements.

\begin{lstlisting}[language=Python]
from PyXAB.synthetic_obj.Garland import Garland
from PyXAB.algos.HCT import HCT
from PyXAB.partition.BinaryPartition import BinaryPartition

# Define the number of rounds, target, domain, partition, and algorithm
T = 1000
target = Garland()
domain = [[0, 1]]
partition = BinaryPartition
algo = HCT(domain=domain, partition=partition)

# Run the algorithm HCT
for t in range(1, T+1):
    point = algo.pull(t)
    reward = target.f(point)
    algo.receive_reward(t, reward)
\end{lstlisting}

%% file: sections/03.Code_Quality.tex
\section{Code Quality and Documentations}
\label{sec: code_quality}

In order to ensure high code quality, we follow the \texttt{PEP8} style and format all of our code using the \texttt{black} package\footnote{\url{https://github.com/psf/black}}. We use the \texttt{pytest} package to test our implementations with different corner cases. More than 95\% of our code is covered by the tests and Github workflows automatically generate a coverage report upon each push or pull request on the main branch\footnote{\url{https://github.com/WilliamLwj/PyXAB}}.

We provide thorough API documentation for each of the implemented classes and functions through numpy docstrings. The documentation is fully available online on ReadTheDocs\footnote{\url{https://pyxab.readthedocs.io/}}. On the same website, we also provide installation guides, both elementary and advanced examples of using our package, as well as detailed contributing instructions and new feature implementation examples to encourage future contributions.
 

%% file: sections/04.Conclusions.tex
\section{Conclusions}
\label{sec: conclusions}

In this paper, we introduce an $\mathcal{X}$-armed bandit algorithm library, PyXAB for online blackbox optimization. The library
contains the implementations of more than 10 $\mathcal{X}$-armed bandit algorithms with complete documentations and clear API references. It also provides different synthetic objectives for evaluation of performance and different choices of the hierarchical partition used in the optimization process. The library serves as the first comprehensive package in the field with code quality assurance to foster future research and generate fair experimental comparisons.